\documentclass[11pt]{article}

\usepackage[final]{acl}
\usepackage{booktabs} 
\usepackage{times}
\usepackage{latexsym}
\usepackage{amsmath}
\usepackage[frozencache=true]{minted}
\usepackage{hyphenat} 
\usepackage{tcolorbox}
\usepackage{listings}

\definecolor{codegreen}{rgb}{0,0.6,0}
\definecolor{codegray}{rgb}{0.5,0.5,0.3}
\definecolor{codepurple}{rgb}{0.58,0,0.82}
\definecolor{backcolour}{rgb}{0.95,0.95,0.92}
\lstdefinestyle{mystyle}{
    backgroundcolor=\color{backcolour},   
    commentstyle=\color{codegreen},
    keywordstyle=\color{magenta},
    numberstyle=\tiny\color{codegray},
    stringstyle=\color{codepurple},
    basicstyle=\ttfamily\footnotesize,
    breakatwhitespace=false,         
    breaklines=true,                 
    captionpos=b,                    
    keepspaces=true,                 
    numbers=left,                    
    numbersep=5pt,                  
    showspaces=false,                
    showstringspaces=false,
    showtabs=false,                  
    tabsize=2
}
\usepackage{cuted}
\usepackage{listings}
\definecolor{bggray}{rgb}{0.95,0.95,0.95}
\usepackage[T1]{fontenc}

\usepackage[utf8]{inputenc}

\usepackage{microtype}

\usepackage{inconsolata}

\usepackage{graphicx}

\title{MeXtract: Light-Weight Metadata Extraction from Scientific Papers}

\author{
Zaid Alyafeai\textsuperscript{1} \,\,\,\,
Maged S. Al-Shaibani\textsuperscript{2} \,\,\,\,
Bernard Ghanem\textsuperscript{1} \\
\textsuperscript{1}KAUST \quad
\textsuperscript{2}SDAIA-KFUPM Joint Research Center for AI, KFUPM
}

\begin{document}
\maketitle
\begin{abstract}
Metadata plays a critical role in indexing, documenting, and analyzing scientific literature, yet extracting it accurately and efficiently remains a challenging task. Traditional approaches often rely on rule-based or task-specific models, which struggle to generalize across domains and schema variations. In this paper, we present MeXtract, a family of lightweight language models designed for metadata extraction from scientific papers. The models, ranging from 0.5B to 3B parameters, are built by fine-tuning Qwen 2.5 counterparts. In their size family, MeXtract achieves state-of-the-art performance on metadata extraction on the MOLE benchmark. To further support evaluation, we extend the MOLE benchmark to incorporate model-specific metadata, providing an out-of-domain challenging subset. Our experiments show that fine-tuning on a given schema not only yields high accuracy but also transfers effectively to unseen schemas, demonstrating the robustness and adaptability of our approach. We release all the \href{https://github.com/IVUL-KAUST/MeXtract}{code}, \href{https://huggingface.co/collections/IVUL-KAUST/mextract-datasets-68e640976c607fc5578f7ed8}{datasets}, and \href{https://huggingface.co/collections/IVUL-KAUST/mextract-68e63fb946b06f5031d4e3ef}{models} openly for the research community.

\end{abstract}

\section{Introduction}

Metadata is an important helper file, usually associated with complex data formats. It usually describes the files using JSON formats with keys and values. Such metadata is essential for indexing data and allowing a simple search mechanism. One of the most important metadata is the one associated with datasets and models. One of the early efforts was Masader \cite{masader}, where the authors collected metadata for hundreds of papers. Such metadata is important to preserve datasets through indexing and searching. It also helps research to thrive by allowing authors to discover datasets related to their work. Similarly, model metadata is an important aspect that allows researchers to track models released and their specific properties. Data and model archives like HuggingFace rely on authors annotating their metadata through data cards. However, this effort is constrained by authors adding their own data description without following a certain format and only using free-form text. 

MOLE \cite{mole} constructed a benchmark and an evaluation framework for evaluating datasets' metadata extracted from scientific papers. It contains 126 arXiv papers annotated using more than 30 attributes for different language categories like Arabic, English, Russian, French, Japanese, and multilingual datasets. However, such an effort is limited, as it relies on zero-shot evaluation of current state-of-the-art models, which are expensive to run. Additionally, the benchmark only considers datasets' metadata.

In this paper, we investigate the ability of light-weight LLMs to extract metadata from long contexts such as scientific papers. Our main contributions are summarized as follows:

\begin{enumerate}
    \item We extend the MOLE benchmark by manually annotating 21 papers for model metadata extraction. 
    \item We introduce an approach for improving metadata extraction through finetuning and preference optimization on a dataset acquired by distilling the knowledge from Kimi k2 \cite{kimi-k2}. 
    \item We release 3 light models named MeXtract, ranging from 0.5B to 3B for efficient metadata extraction. Our experiments show that such models achieve state-of-the-art results on metadata extraction compared to similarly sized models. 
\end{enumerate}

This research results in multiple artifacts, including 3 datasets for resource evaluation, instruction tuning, and preference optimization. In addition, we release all models ranging from 0.5B to 3B using the Apache-2.0 license to advance research in this domain. We will release all the datasets and models in the non-anonymous version of the paper. 

\section{Related Work}

Controlled output generation has been gaining a lot of interest recently. As a result, many tools have been developed to support structured output, like: XGrammar \cite{xgrammar}, Outlines \cite{outlines}, and SGLang \cite{sglang}. SciTreck \cite{li2025gets} was introduced to evaluate LLMs' long-context reasoning capabilities on a benchmark of scientific articles. However, such approaches could affect performance, especially with complex schema \cite{schemabench}.

To improve performance for controlled output generation, there have been some studies that used schema-based generation to generate better output. For example, \cite{schemabench} introduced SchemaBench to evaluate LLMs' ability to output constraint schema. They use reinforcement learning rather than instruction tuning to improve performance. Recently, there have been many attempts to generalize that to extracting metadata from scientific papers. MOLE \cite{mole} introduced a benchmark and a framework for evaluating the capabilities of LLMs to extract metadata from scientific papers. They mainly analyze the abilities of flagship LLMs through a schema-based approach. They also analyzed different properties for extraction, like browsing, few-shot, and varying context length. \cite{watanabeCapabilitiesChallengesLLMs2025} evaluated the capabilities of two LLMs, namely Misteral 7B and Llama 3 8B, to extract metadata from scientific papers. They use 777 papers from the proceedings of the main track of ACL 2020. They show that such LLMs are not a feasible approach for metadata extraction, as the accuracy is not necessarily high.  \cite{giner-miguelezUsingLargeLanguage2024} compared the abilities of two LLMs, namely GPT-3 and Flan-UL2, to extract metadata from 12 scientific dataset papers. They utilize a chain of prompts to answer questions about the dataset's metadata, given a context extracted using a retrieval model. 

Recently, there were many specialized tools for the extraction of general metadata from any given text. NuExtract \cite{nuextract} introduced 3 models, 2B, 4B, and 8B, to extract metadata using the vision capabilities of Qwen2.5-VL. They use a schema-based approach to extract the LLMs. However, they evaluate their approach using a collection of examples that are not necessarily long context. Additionally, they rely mostly on simple metadata extraction on a few attributes. Similarly, LangExtract \cite{langextract} is a tool for metadata extraction introduced by Google. It requires precise control from the user to extract metadata from unstructured text. Additionally, the user has to create sample examples to extract the structured output correctly. Additionally, it is quite slow compared to predicting directly from an LLM. 

Compared to other approaches, MeXtract models' capabilities are demonstrated on long context using scientific papers. Additionally, they don't require using examples to extract metadata efficiently. We also open-source the finetuning and evaluation data for the open research community.  

\section{Background}
We define metadata as a JSON with a set of keys and values $\{A_n\} \to \{V_n\}$. Each key represents the name of the attribute, and its value is the extracted value of such an attribute. Each extracted metadata is constrained by a schema that defines three variables, namely, Type, Length, and Options. The type defines the return type of the attribute. For example, the age of a person must be an integer. Length defines the minimum and maximum range of a given value. For example, we could constrain the Year attribute to be between 2000 and 2025. This attribute is attached to the type; for example, if the type is a list, then it defines the number of item values to output. On the other hand, if it is a string, it defines the number of words in the answer. Finally, certain attributes might define options, which are a finite set of values to choose from. Any given extraction method must extract the values from the options only. Here is an example schema for an attribute\footnote{In the Appendix, we include a complete schema for multiple attributes. }:

\begin{minted}[frame=single,fontsize=\small,breaklines]{json}
{
    "Unit": {
        "answer_type": "str",
        "answer_min": 1,
        "answer_max": 1,
        "options": [
            "Million",
            "Billion",
            "Trillion"
        ]
    }
}
\end{minted}

An extracted metadata is considered valid if it returns the correct type for each attribute. An extraction method can return the wrong answer min or answer max of a given attribute. We make this relaxation because current LLMs struggle to follow exact length constraints \cite{s1}. An incorrect type can be cast to the correct type. For example, if the Version of a model has a \texttt{Float} type but the value returned is a string, then we can cast it like \texttt{float(value)} to return the correct type. The casting mechanism can be implemented easily using Python. 

In our setup, we want to extract metadata from a source paper text. The output metadata is defined by a JSON that can be read using Python's \texttt{json.loads(metadata)}. Our extracted metadata is mostly related to deep learning artifacts, which are datasets and models. However, the available benchmark MOLE \cite{mole} only considers datasets. Hence, we extend this benchmark to include metadata for models. 

\section{MOLE+ Benchmark}
The MOLE \cite{mole} benchmark consists of 126 arXiv datasets' papers annotated with around 30 attributes. It covers six different categories: Arabic, English, Russian, Japanese, French, and Multilingual datasets. Each category contains exactly 21 papers. We extend the benchmark to include metadata for models. We manually collect 21 papers of the most popular models on the HuggingFace Hub\footnote{HuggingFace Hub: \url{https://huggingface.co/models}}. We collect a variety of models spanning different modalities like text, image, and audio. We also consider different architectures like transformers, recurrent neural networks (RNNs), and state space models (SSMs). To annotate the papers, we use a semi-automated method where we first use Kimi K2 \cite{kimi-k2} to annotate the metadata, and then review them manually. The model schema contains 16 attributes defined as follows:

\begin{figure*}
\centering
\includegraphics[width = 0.9\linewidth]{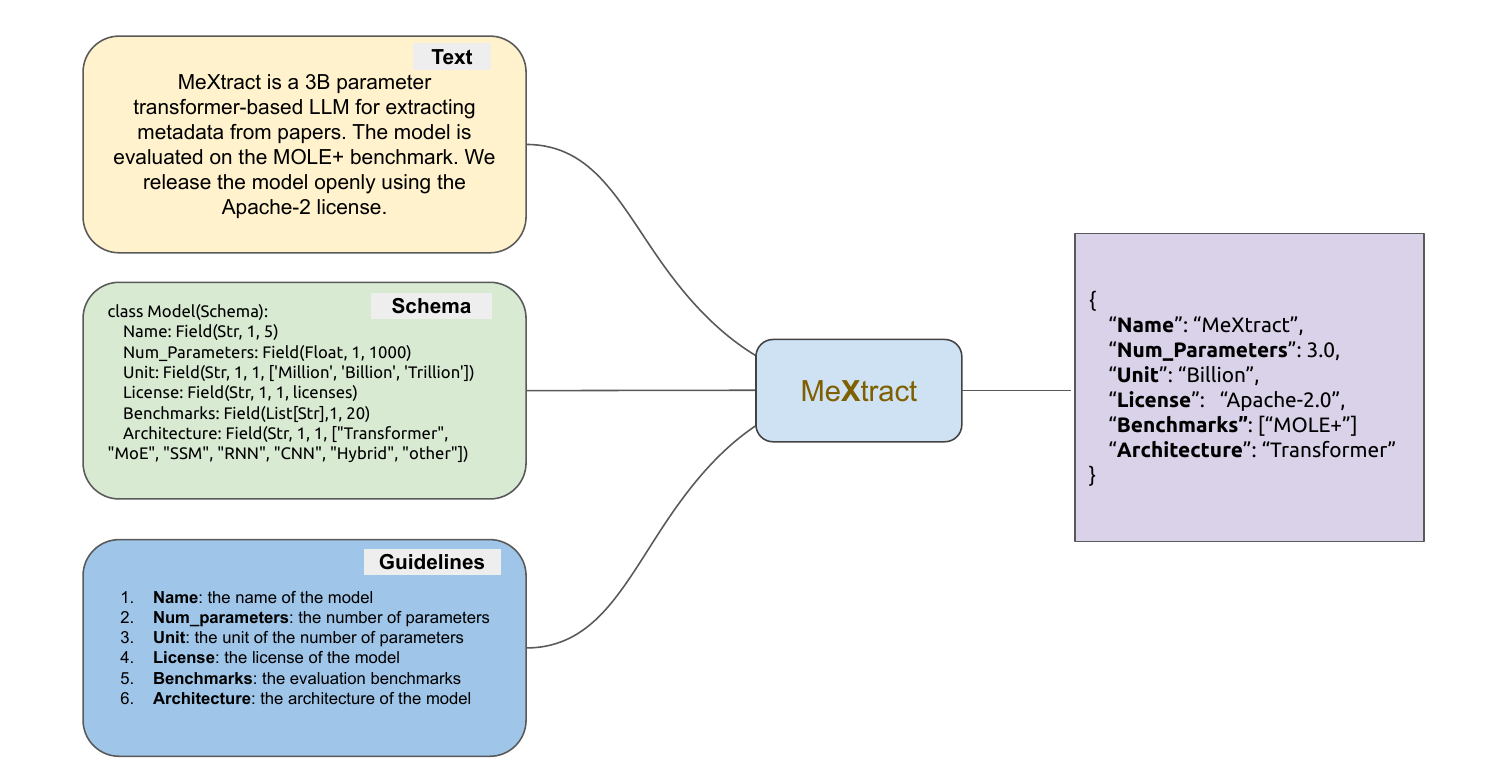}
\caption{Schema-based Metadata extraction using MeXtract. The model has three inputs: the paper text, schema, and guidelines, and one output, which is the metadata JSON. This example is only for illustration purposes; the model schema contains 16 attributes.}
\label{fig:method}
\end{figure*}

\begin{enumerate}
    \item \textbf{Name} Name of the model family.
    \item \textbf{Num Parameters} The number of parameters of the largest model. If there are multiple models in the paper, choose the biggest one (mostly the chat version).
    \item \textbf{Unit} The unit of the number of parameters. We choose from "Million", "Billion", and "Trillion" of parameters.
    \item \textbf{Type} The type of the model. The model type could be base, chat, or code. 
    \item \textbf{Think} Whether the model is a thinking/reasoning model or not.
    \item \textbf{Version} The version of the model. Namely, a float value.
    \item \textbf{Models} All model variants listed in the paper are defined using a dictionary.
    \item \textbf{License} The license of the model.
    \item \textbf{Year} The year the model was published.
    \item \textbf{Benchmarks} What benchmarks were the model evaluated on?
    \item \textbf{Architecture} The architecture of the model.
    \item \textbf{Context} The context size of the model.
    \item \textbf{Language} The type of pretraining dataset for the model. It could be monolingual, bilingual, or multilingual. 
    \item \textbf{Provider} The company or organization that provided the model.
    \item \textbf{Modality} The modality of the model. 
    \item \textbf{Paper Link} The link to the paper.
\end{enumerate}

Similar to MOLE, each attribute in the schema will have a binary value of 1 if it can be extracted from the paper and 0 otherwise\footnote{This helps us in evaluating the precision and recall.}. Hence, some attributes might not exist in the paper. As an example, the License attribute might exist on an external repository like GitHub or HuggingFace. We create this model schema as an out-of-distribution test dataset, where most of the attributes don't exist in the original MOLE benchmark. We want to measure the models' ability to generalize to unseen model schemas.

For model inference and training, we pass the schema in addition to the annotation guidelines for the model to predict the output JSON. The annotation guidelines are the same ones used to manually annotate the data. Check Figure \ref{fig:method} for an overview of our approach. We distinct this from \cite{mole}, where the authors passed the guidelines and schema as one file. 

Below, we show an example of annotated metadata for a given paper:

\begin{minted}[frame=single,fontsize=\small,breaklines]{json}
{
    "Name": "BERT",
    "Num_Parameters": 340.0,
    "Unit": "Million",
    "Type": "Base",
    "Think": false,
    "Version": 1.0,
    "Models": [
        {
            "Name": "BERT_BASE",
            "Num_Parameters": 110,
            "Unit": "Million",
            "Type": "Base",
            "Think": false
        },
        {
            "Name": "BERT_LARGE",
            "Num_Parameters": 340,
            "Unit": "Million",
            "Type": "Base",
            "Think": false
        }
    ],
    "License": "Apache-2.0",
    "Year": 2018,
    "Benchmarks": [
        "GLUE",
        "SQuAD v1.1",
        "SQuAD v2.0",
        "SWAG",
        "CoNLL-2003 NER"
    ],
    "Architecture": "Transformer",
    "Context": 512,
    "Language": "monolingual",
    "Provider": "Google AI",
    "Modality": "text",
    "Paper_Link": "https://arxiv.org/pdf/1810.04805"
}
\end{minted}

\section{Methodology}
In this section, we highlight our methodology, with details of preprocessing, instruction tuning, preference optimization, and evaluation.  

\subsection{Preprocessing}
We use a schema-based approach following the work of \cite{nuextract} and \cite{mole} with some modifications. In addition to the text, we pass two extra inputs: 1) Schema, which defines the types, length, and options if they exist, and 2) Guidelines, which define each attribute in the text. We differentiate our work from MOLE \cite{mole}, where the schema and guidelines are mixed together as one input. Additionally, this makes our work generic, as any new schema can be defined. Furthermore, we make a resemblance to human annotation where the annotators are provided guidelines to extract annotated text from a given corpus. For a given LLM, we convert the Schema to a JSON text format to be passed as input. For the paper text, we use PDF Plumber\footnote{\url{https://github.com/jsvine/pdfplumber}} from Python to convert the given PDF to text format. Although some papers might have source files using LaTeX, we want our approach to be general and robust against noisy text. We use Qwen2.5 Instruct \cite{qwen2.5} as our base model for finetuning. For each model, we truncate the text to fit in a model length of 8192 tokens. 

\subsection{Supervised Instruction Tuning}

\begin{figure*}[!h]
    \centering
    \includegraphics[width=\linewidth]{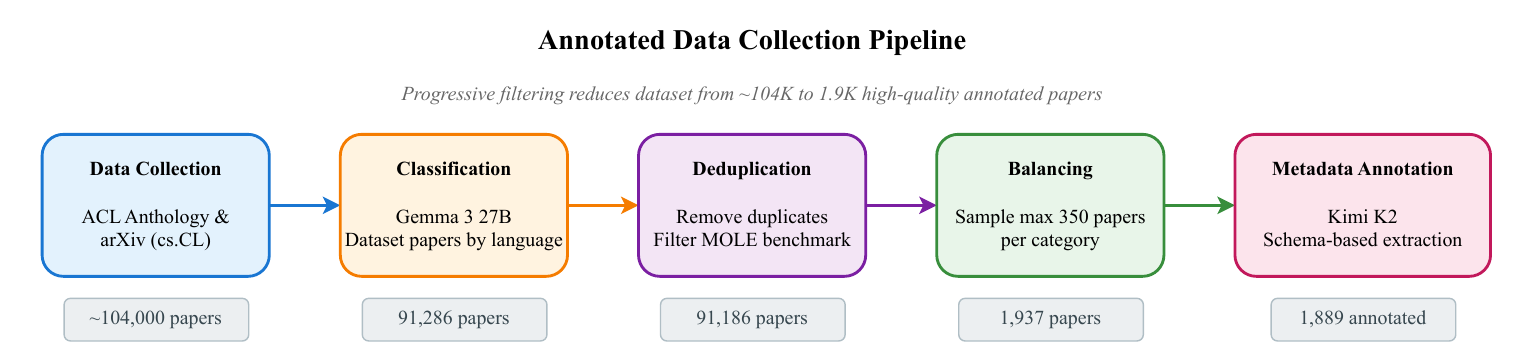}
    \caption{Annotated data collection pipeline for instruction tuning. In each stage, we show the number of papers. }
    \label{fig:data_collection}
\end{figure*}

We create a dataset of 1,889 annotated papers to finetune our models. The dataset is used to create the SFT models. We rely on 5 steps to collect the SFT dataset as illustrated in Figure \ref{fig:data_collection}. These steps are detailed below.

\begin{enumerate}
    \item \textbf{Data Collection} We collect around 104K papers from the ACL Anthology and arXiv, filtering arXiv to only include datasets from the \texttt{cs.CL} category. For each paper, we extract the title and abstract. 
    \item \textbf{Classification} In this stage, we are interested to classify dataset papers. This includes papers that introduce a new dataset. We use Gemma 3 27B to classify the datasets into seven categories: \texttt{ar}, \texttt{en}, \texttt{ru}, \texttt{fr}, \texttt{jp}, \texttt{multi}, \texttt{other}, and \texttt{none}. The \texttt{other} category contains resource papers not covered by the mentioned languages. While the \texttt{none} category include papers that are not resource papers. We highlight the system prompt in Figure \ref{fig:system_prompt}.
    \item \textbf{Deduplication} We use the title and abstract to identify duplicate papers. Duplicates mostly exist if a paper is published both in arXiv and the ACL Anthology. After removing duplicates and keeping only correctly classified papers, we obtain 91,286 papers. We also apply a second phase of deduplication to remove papers that exist in the MOLE benchmark. This reduces the size of papers to around 91,186.
    \item \textbf{Balancing} We balance the labels by extracting categories with at most 350 papers, except for the Russian split, which contains 187 papers. This results in 1,937 papers across six categories.
    \item \textbf{Metadata Annotation} We use Kimi-K2 \cite{kimi-k2} to annotate metadata for the 1,937 papers. Our experiments show that Kimi-K2 is competitive in metadata extraction while being open source (see Table \ref{tab:flagship}). We use the same approach as in Figure \ref{fig:method} to extract metadata for all the papers. After skipping errors, we are left with 1,889 annotated papers. 
    
\end{enumerate}


We collected around 1,889 annotated data points using synthetic generation, which we split into 95\% for training and 5\% for validation with a fixed random seed for reproducibility. We finetune Qwen 2.5 models at three scales (0.5B, 1.5B, and 3B parameters) using parameter-efficient LoRA \cite{hu2022lora} finetuning to create the MeXtract model family. 

For LoRA configuration, we use rank $r=8$ and $\alpha=16$. We train for 10 epochs using the AdamW \cite{loshchilov2017decoupled} optimizer with a learning rate of $2 \times 10^{-4}$, linear learning rate scheduling with 5 warmup steps, and weight decay of 0.01. The effective batch size is 8 (batch size of 2 with 4 gradient accumulation). We set the maximum context length to 8,192 tokens with 2,048 tokens reserved for outputs. Training uses \texttt{bfloat16} precision and applies loss only on the completion tokens rather than the full sequence. We implement early stopping with a patience of 10 evaluation steps and a minimum improvement threshold of 0.001, monitoring validation loss to select the best checkpoint.

\subsection{Preference Optimization}
To further improve the model, we use direct preference optimization (DPO) \cite{dpo}. We first filter the SFT data to only include the metadata that follows the \textit{Length} constraint. This stage resulted in around 1,174 data samples. Then we apply a set of transformations to create the rejected samples: 

\begin{enumerate}
    \item \textbf{Malformed JSON} We create incorrect JSON examples by randomly stripping, converting double quotation marks to single quotation marks, and removing commas. 

    \item \textbf{Different Format} we add the \texttt{answer} key to some examples. Also, we convert some metadata from JSON to Markdown format. These heuristics are created based on inspecting the issues with some previous inference runs. 
    
    \item \textbf{Modify the length} We randomly modify the lengths of some attributes to not follow the length constraints. For example, we change the Year to be outside the specified range. Additionally, for answers with list output, we add more options to be larger than the specified min and max. 
\end{enumerate}

We split this dataset into 939 for training and 235 for validation. For finetuning, we use the \texttt{trl}\footnote{\url{https://github.com/huggingface/trl}} library for preference optimization. We use a learning rate of $5.0\times 10^{-6}$ and a batch size of 16 for 300 steps. We evaluate every 100 steps. We use a LoRA with rank 8 and alpha 16. All the experiments are run on a single A100 with around 80GB of GPU memory. 

\begin{table*}[!htp]
\centering
\caption{Model performance across different categories. We group zero-shot models vs. schema-based approaches compared to our MeXtract models. }
\label{tab:model-results}
\begin{tabular}{lcccccccc}
\toprule
\textbf{Model}                & \textbf{ar} & \textbf{en} & \textbf{jp} & \textbf{fr} & \textbf{ru} & \textbf{multi} & \textbf{model} & \textbf{Average} \\
\midrule
\textbf{Falcon3 3B Instruct}  & 20.46 & 16.30 & 20.29 & 17.81 & 17.23 & 16.13 & 15.96 & 17.74 \\
\textbf{Llama3.2 3B Instruct} & 28.77 & 25.17 & 33.14 & 27.73 & 22.21 & 22.58 & 33.37 & 27.57 \\
\textbf{Gemma 3 4B It}        & 44.88 & 46.50 & 48.46 & 43.85 & 46.06 & 42.05 & 56.04 & 46.83 \\
\textbf{Qwen2.5 3B Instruct}  & 49.99 & 56.72 & 61.13 & 57.08 & 64.10 & 52.07 & 59.05 & 57.16 \\
\hline
\textbf{MOLE 3B}                 & 23.03 & 50.88 & 50.83 & 50.05 & 57.72 & 43.34 & 17.17 & 41.86 \\
\textbf{Nuextract 2.0 4B}     & 44.61 & 43.57 & 43.82 & 48.96 & 47.78 & 40.14 & 49.90 & 45.54 \\
\textbf{Nuextract 2.0 8B}     & 51.93 & 58.93 & 62.11 & 58.41 & 63.21 & 38.21 & 53.70 & 55.21 \\
\hline
\textbf{MeXtract 0.5B}    & 65.96 & 69.95 & 73.79 & 68.42 & 72.07 & 68.20 & 32.41 & 64.40 \\
\textbf{MeXtract 1.5B}    & 67.06 & 73.71 & 75.08 & 71.57 & 76.28 & 71.87 & 52.05 & 69.66 \\
\textbf{MeXtract 3B}      & \textbf{70.81} & \textbf{78.02} & \textbf{78.32} & \textbf{72.87} & \textbf{77.51} & \textbf{74.92} & \textbf{60.18} & \textbf{73.23} \\
\bottomrule
\end{tabular}
\end{table*}

\subsection{Evaluation}
We split our evaluation strategy into two main strategies. 

\textbf{Performance Evaluation}: In this strategy, we evaluate the extracted metadata by comparing the predicted values with the gold values. Compared to MOLE \cite{mole}, where they used a simple approach for evaluation, relying on exact match, which penalizes models that might create similar output. For example, the following instance \texttt{exact\_match('SQuAD 1.0', 'SQuAD v1.0')} will result in zero when evaluated using the MOLE approach. To circumvent this, we use validated types to correctly evaluate all the attributes. Here is a list of all data types in our schema with their corresponding comparison approach:

\begin{itemize}
    \item \texttt{Numbers:} for values that are numbers like integers or floats, we use the following formula: $$1- \frac{|X-Y|}{\text{max}(X, Y)}$$
    where $X$ represents the gold value, and $Y$ represents the predicted value. For the Year attribute, we use the same formula after subtracting 2010. 
    \item \texttt{String}: we use Edit Distance if there are no options, otherwise we use exact match. 
    \item \texttt{LongStr:} this is used for long outputs like abstract, description, etc. We use the Rouge score for such an attribute. 
    \item \texttt{URL:} like short strings, we use the Edit Distance. 
    \item \texttt{List:} we use the following formula:
    $$\frac{X \cap Y}{\text{max}(X, Y)}$$
    This formula calculates the intersection between the two lists, normalized by the maximum length. 
\end{itemize}

Similar to \cite{mole}, we then calculate the precision and recall for all attributes. The precision calculates the match between predicted attributes and gold attributes, while recall only considers attributes that exist in the paper. We then calculate the F1 score as the final result for a given prediction.


\textbf{Constraints Following Evaluation} Every attribute in the schema is annotated with length constraints. In this metric, we evaluate the model's ability to follow the length constraints defined in the schema. If the attribute lies within the range defined in the length attribute, it is given the score 1; otherwise, it is 0. The length attribute is attached to the type and it is defined using \texttt{answer\_min} and \texttt{answer\_max}. Such inputs are given to the model and are expected to follow such constraints. For example, for the \texttt{License} attribute, we have a range of \texttt{(1,1)}, which means we must output only one item from the list of options. 
\begin{table}[!htp]
\centering
\caption{Number of parameters and context length for selected models.}
\label{tab:model_context}
\begin{tabular}{lcc}
\toprule
\textbf{Model} & \textbf{Parameters} & \textbf{Context} \\
\midrule
\textbf{Falcon 3} & 3B & 32K tokens \\
\textbf{Qwen 2.5} & 3B & 128K tokens \\
\textbf{LlaMA 3.2} & 3B & 128K tokens \\
\textbf{Gemma 3} & 4B & 128K tokens \\
\bottomrule
\end{tabular}
\end{table}
\section{Results and Discussions}
We group our models for evaluation into two categories: 1) Zero-shot LLMs, similar model families, namely the instruct versions of the following models: Gemma3 4B \cite{gemma3}, Qwen2.5 3B \cite{qwen2.5}, Falcon3 3B \cite{falcon3}, and Llama 3.2 3B \cite{llama3} (see Table \ref{tab:model_context}) and 2) Metadata extraction methods, namely MOLE \cite{mole} and NuExtract\cite{nuextract}. For MOLE, we use the same schema and Qwen2.5 3B Instruct to predict the metadata. For each model, we use the default inference parameters to be more consistent. For each inference, we try a maximum of three attempts; if the model fails, we use a default empty metadata as output. 

\subsection{Performance Evaluation}
In Table \ref{tab:model-results}, we compare the results of all of our models to the 7 other models. We observe that MeXtract 3B achieves the highest results across all categories. Remarkably, MeXtract 0.5B and 1.5B models achieve the highest results in all categories except for the model category. Notably, our 3B model generalizes well to unseen schemas on the model category. We note some models, like Falcon 3, fail at following the schema, resulting in almost random results. Interestingly, Qwen2.5 3B has strong zero-shot abilities, resulting in better results compared to schema-based approaches, MOLE, and finetuned models like Nuextract. In summary, our 3B model achieves state-of-the-art results, even beating models that are larger in size. 

\begin{figure*}[!htp]
    \centering
    \includegraphics[width=0.85\linewidth]{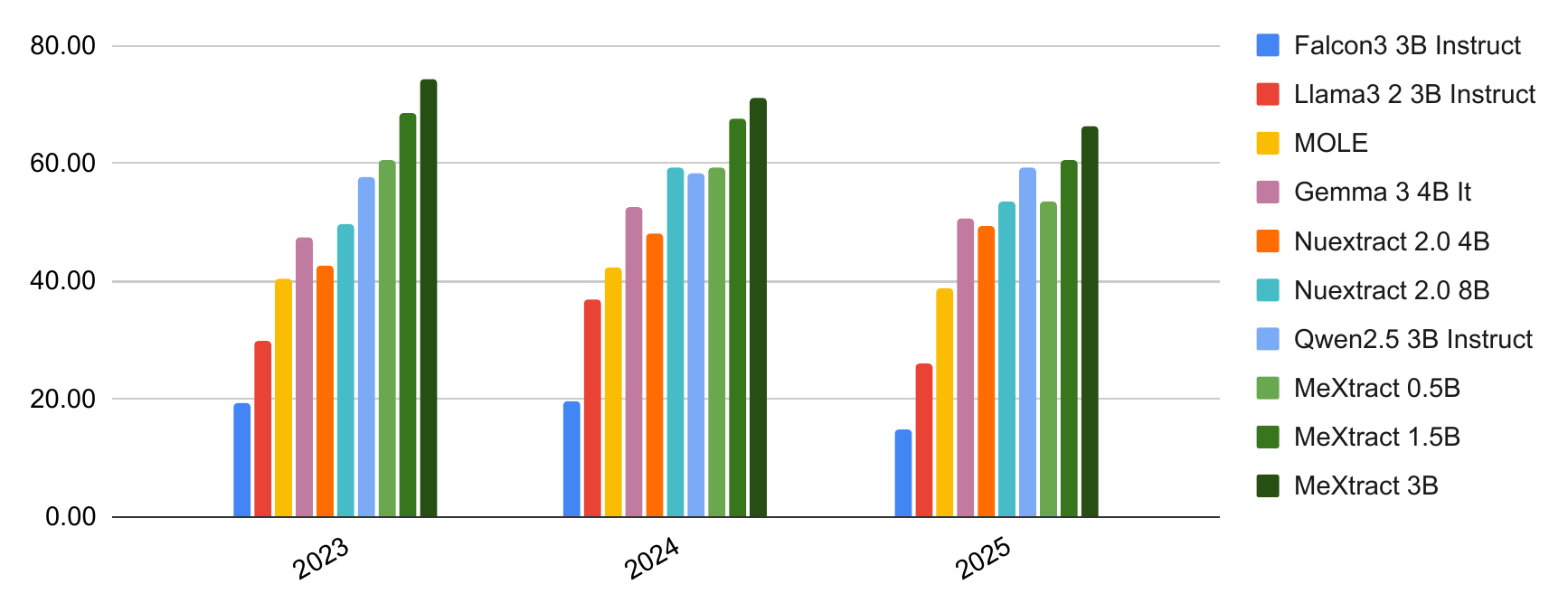}
    \caption{Results of all models averaged by year. The results show the F1 score for the years 2023, 2024, and 2025, respectively. }
    \label{fig:year}
\end{figure*}

\subsection{Constraints Following Evaluation}
In this experiment, we want to measure our model's ability to follow the length constraints compared to other models. In Table \ref{tab:length}, we show the results of all models in terms of length averaged across all categories in MOLE+. We observe that all of our models achieve the highest results in terms of following the length constraints. Also, increasing the model size results in better constraint following. We want to highlight that models suffering from errors might have inflated scores because the default JSON results in outputs with relaxed constraints. 
\begin{table}[h!]
\centering
\caption{Model performance comparison using the length constraint. The length is calculated as the average of all schema categories in MOLE+.}
\label{tab:length}
\begin{tabular}{lc}
\toprule
\textbf{Model} & \textbf{Length} \\
\midrule
\textbf{Falcon3 3B Instruct}  & 86.88 \\
\textbf{Llama3.2 3B Instruct} & 90.22 \\
\textbf{MOLE 3B}                 & 91.42 \\
\textbf{Nuextract 2.0 4B}     & 96.20 \\
\textbf{Gemma 3 4B It}        & 96.61 \\
\textbf{Nuextract 2.0 8B}     & 97.11 \\
\textbf{Qwen2.5 3B Instruct}  & 97.67 \\
\textbf{MeXtract 0.5B}        & 97.93 \\
\textbf{MeXtract 1.5B}        & 98.81 \\
\textbf{MeXtract 3B}          & \textbf{99.24} \\
\bottomrule
\end{tabular}
\end{table}
\subsection{Preference optimization}
In this experiment, we want to observe the effect of using preference optimization vs. just instruction tuning. We train two models for each size, one with SFT only and one with SFT and DPO. In Table \ref{tab:mextract-dpo}, we show the results with and without DPO. We see a clear advantage across all models using DPO. The advantage is smaller for the 3B model compared to the smaller models. These results highlight the importance of using preference optimization to align LLMs to follow constrained output.  
\begin{table}[ht]
\centering
\caption{Average performance comparison of MeXtract models with SFT Only and using SFT+DPO. The results are shown as the average of all categories in MOLE+.}
\label{tab:mextract-dpo}
\begin{tabular}{lrr}
\toprule
\textbf{Model Size} & \textbf{+SFT} & \textbf{+DPO} \\
\midrule
\textbf{MeXtract 0.5B} & 63.69 & \textbf{64.40} \\
\textbf{MeXtract 1.5B} & 68.63 & \textbf{69.66} \\
\textbf{MeXtract 3B}   & 73.06 & \textbf{73.23} \\
\bottomrule
\end{tabular}
\end{table}
\begin{figure}[!htp]
    \centering
    \includegraphics[width=\linewidth]{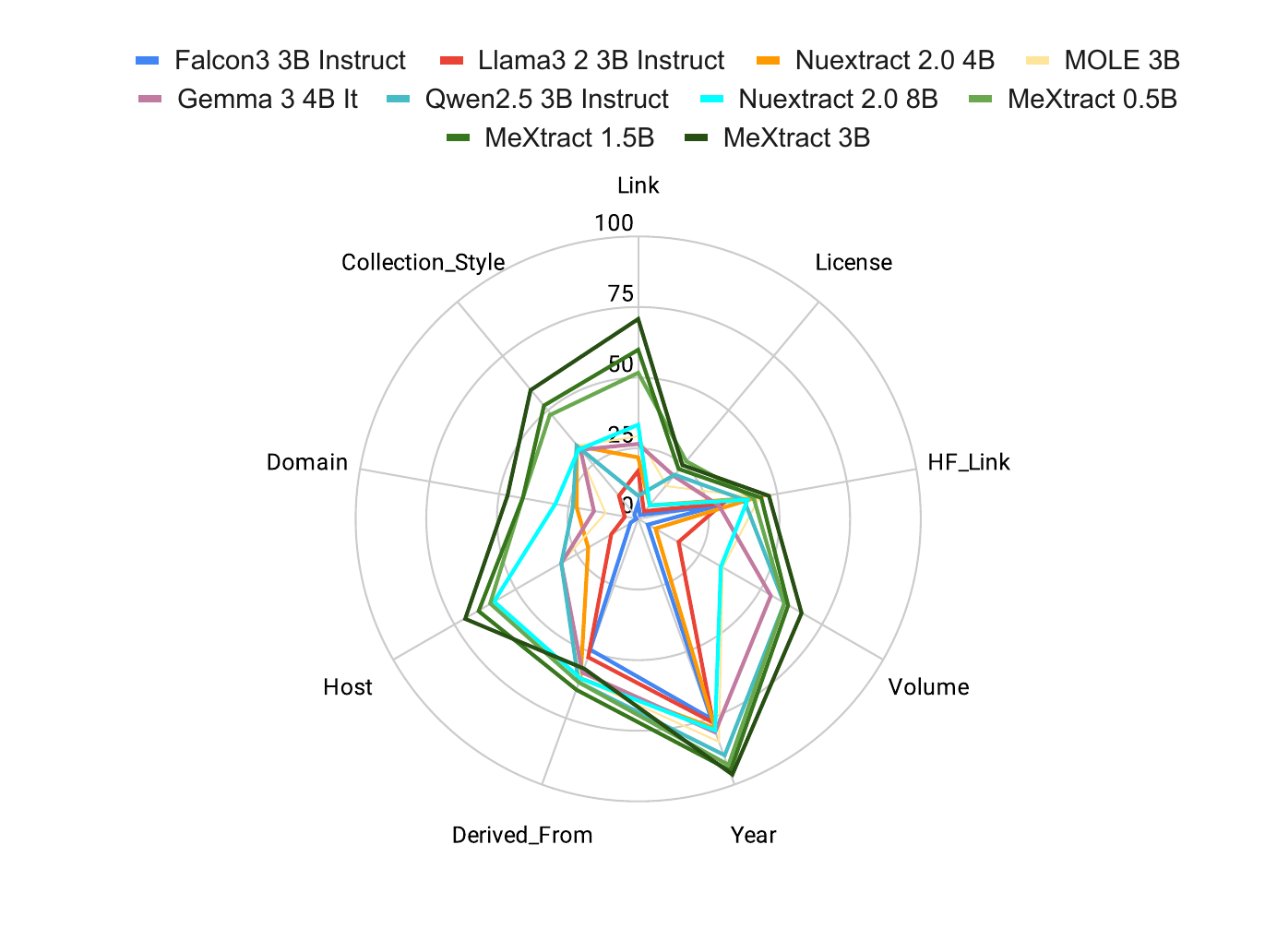}
    \caption{Results per 9 attributes for all the models using the MOLE benchmark.}
    \label{fig:radar}
\end{figure}

\subsection{Individual Attributes}
In Figure \ref{fig:radar} we show the results for some attributes using the MOLE benchmark. We highlight that for some attributes, it might be easier for the models to extract. For example, the Year and Link might be much easier to extract because such attributes are clearly defined in published papers. Some attributes might not exist in the paper, like the License, so it will prove to be more difficult (much lower precision compared to recall). Still, our models achieve the highest score, with the 3B model achieving the highest result in 7 out of 9 attributes. 

\subsection{Contamination}
Since our pretrained models are based on Qwen2.5, which uses the web as pretraining data, such models might have been exposed to scientific literature. This might result in contamination during pre-training. To observe that, we only evaluate on papers released after 2025 and compare them to papers released before. In Figure \ref{fig:year}, we show the results of all models in each year, ranging from 2023 to 2025. Mainly, we observe that Qwen2.5 3B doesn't suffer from contamination as the results show an average result of around 58\% for all the years. We show the results for the years from 2014 to 2022 in MOLE+ in Table \ref{tab:by_year_old}. 

\section{Conclusion}
In this paper, we introduce a set of light-weight metadata extraction models called MeXtract. The models range from 0.5B to 3B based on finetuning Qwen 2.5 counterpart models. The models are finetuned using synthetic instruction data, followed by reinforcement learning learning through preference optimization. To evaluate the models, we first extend the MOLE benchmark to include model metadata. Through multiple experiments, we show that our models achieve state-of-the-art results for extracting metadata from scientific papers in their size family. Additionally, we show that our models extend to unseen metadata by evaluating them on the model schemas that were unseen during finetuning. 

\section*{Limitations}
In this section, we summarize the limitations of our work:
\begin{itemize}
    \item \textbf{Input Format} We mainly rely on text extracted from PDFs to be passed to our models. This might not generalize well to out-of-domain input formats. Nevertheless, we consider this input format as a noisy text because PDF extractors use traditional approaches.    

    \item \textbf{Nested Metadata} our approach doesn't measure complex metadata where there could be multiple nested variables. We tested only with first-level nested metadata, but it will be worthwhile to consider even second and third-level nested attributes. 

    \item \textbf{Limited performance compared to larger LLMs} In this paper, we only consider lightweight models that are small in size as fast metadata extractors. Compared to flagship models that could have +100 billion parameters, our models have yet to achieve comparable performance (see Table \ref{tab:flagship}). We think our approach can still be generalized to achieve comparable performance by collecting more data. 
    
\end{itemize}

\section*{Acknowledgements}
The research reported in this publication was sup-
ported by funding from King Abdullah University
of Science and Technology (KAUST) - Center of
Excellence for Generative AI, under award number
5940.
\bibliography{custom}

\appendix

\section{Models}
\label{sec:models}
In Table \ref{tab:license}, we summarize all the models used in this paper and their links and licenses. All the models can be used for research purposes. 

\section{Metrics}
In Table \ref{tab:metrics}, we show the results for precision, recall, and F1 scores. In general, we observe that the precision is always lower because the model might hallucinate a lot of attributes, but the recall is higher because it only evaluates on attributes that can be extracted from the paper. 

\section{Evaluation By Year}
In Table \ref{tab:by_year_old}, we show the results of all papers published before 2023. We show that even for older papers, our models achieve the highest results for most of the years. 

\section{Flagship Models}
In Table \ref{tab:flagship}, we compare the results to flagship models in the literature. As expected, larger LLMs are still much better in the retrieval task, especially those that require long context. 

\begin{table}[h!]
\centering
\caption{Precision, recall, and F1 scores of our models compared to much larger models.}
\label{tab:flagship}
\begin{tabular}{lccc}
\toprule
\textbf{Model} & \textbf{Precision} & \textbf{Recall} & \textbf{F1} \\
\midrule
\textbf{MeXtract 0.5B}    & 63.29 & 65.62 & 64.40 \\
\textbf{MeXtract 1.5B}    & 68.45 & 70.98 & 69.66 \\
\textbf{MeXtract 3B}      & 72.02 & 74.56 & 73.23 \\ \hline
\textbf{Gemini 2.5 Pro} & 77.58 & 80.45 & 78.96 \\
\textbf{Kimi K2}    & 79.34 & 82.08 & 80.66 \\
\textbf{Grok 4}           & \textbf{79.87} & \textbf{83.19} & \textbf{81.47} \\
\bottomrule
\end{tabular}
\end{table}

\begin{table*}[!htp]
\centering
\caption{Open models used in this research with their corresponding links and licenses. }
\label{tab:license}
\begin{tabular}{lll}
\toprule
\textbf{Model} & \textbf{Link} & \textbf{License} \\
\midrule
\textbf{Falcon3 3B Instruct} & \href{https://hf.co/tiiuae/Falcon3-3B-Instruct}{tiiuae/Falcon3-3B-Instruct} & TII Falcon-LLM License 2.0 \\
\textbf{Llama3 2 3B Instruct} & \href{https://hf.co/meta-llama/Llama-3.2-3B-Instruct}{meta-llama/Llama-3.2-3B-Instruct} & Llama 3.2 Community License \\
\textbf{Gemma 3 4B It} & \href{https://hf.co/google/gemma-3-4b-it}{google/gemma-3-4b-it} & Gemma \\
\textbf{Gemma 3 27B It} & \href{https://hf.co/google/gemma-3-27b-it}{google/gemma-3-27b-it} & Gemma \\
\textbf{Qwen2.5 0.5B Instruct} & \href{https://hf.co/Qwen/Qwen2.5-0.5B-Instruct}{Qwen/Qwen2.5-0.5B-Instruct} & Apache 2.0 \\
\textbf{Qwen2.5 1.5B Instruct} & \href{https://hf.co/Qwen/Qwen2.5-1.5B-Instruct}{Qwen/Qwen2.5-1.5B-Instruct} & Apache 2.0 \\
\textbf{Qwen2.5 3B Instruct} & \href{https://hf.co/Qwen/Qwen2.5-3B-Instruct}{Qwen/Qwen2.5-3B-Instruct} & Qwen Research License \\
\textbf{Nuextract 2.0 4B} & \href{https://hf.co/numind/NuExtract-2.0-4B}{numind/NuExtract-2.0-4B} & MIT \\
\textbf{Nuextract 2.0 8B} & \href{https://hf.co/numind/NuExtract-2.0-8B}{numind/NuExtract-2.0-8B} & MIT \\
\textbf{Kimi K2} & \href{https://hf.co/moonshotai/Kimi-K2-Instruct}{moonshotai/Kimi-K2-Instruct} & Modified MIT \\
\bottomrule
\end{tabular}
\end{table*}

\begin{table*}[h!]
\centering
\caption{Evaluation results (precision, recall, F1) for different models.}
\label{tab:metrics}
\begin{tabular}{lccc}
\toprule
\textbf{Model} & \textbf{Precision} & \textbf{Recall} & \textbf{F1} \\
\midrule
\textbf{Falcon3 3B Instruct}  & 17.76 & 17.83 & 17.74 \\
\textbf{Llama3 2 3B Instruct} & 27.28 & 27.94 & 27.57 \\
\textbf{Mole}                 & 41.30 & 42.50 & 41.86 \\
\textbf{Nuextract 2.0 4B}     & 44.67 & 46.51 & 45.54 \\
\textbf{Gemma 3 4B It}        & 46.30 & 47.48 & 46.83 \\
\textbf{Nuextract 2.0 8B}     & 54.13 & 56.41 & 55.21 \\
\textbf{Qwen2.5 3B Instruct}  & 56.21 & 58.21 & 57.16 \\
\textbf{MeXtract 0.5B}    & 63.29 & 65.62 & 64.40 \\
\textbf{MeXtract 1.5B}    & 68.45 & 70.98 & 69.66 \\
\textbf{MeXtract 3B}      & \textbf{72.02} & \textbf{74.56} & \textbf{73.23} \\
\bottomrule
\end{tabular}
\end{table*}

\begin{table*}[h!]
\centering
\caption{Performance of models in papers published from 2014–2022.}
\label{tab:by_year_old}
\begin{tabular}{lccccccccc}
\toprule
\textbf{Model} & \textbf{2014} & \textbf{2015} & \textbf{2016} & \textbf{2017} & \textbf{2018} & \textbf{2019} & \textbf{2020} & \textbf{2021} & \textbf{2022} \\
\midrule
\textbf{Falcon3 3B Instruct}  & 20.30 & 15.37 & 17.65 & 25.59 & 13.24 & 14.66 & 18.16 & 17.92 & 16.11 \\
\textbf{Llama3 2 3B Instruct} & 8.40  & 51.81 & 18.01 & 23.84 & 23.81 & 18.32 & 21.59 & 28.38 & 28.33 \\
\textbf{Mole}                 & 23.83 & 22.77 & 36.92 & 46.67 & 44.03 & 40.44 & 44.67 & 46.03 & 41.41 \\
\textbf{Gemma 3 4B It}        & 45.83 & 19.41 & 44.57 & 47.08 & 49.41 & 40.40 & 42.17 & 49.52 & 41.08 \\
\textbf{Nuextract 2.0 4B}     & 51.28 & 43.53 & 41.96 & 42.57 & 46.14 & 41.06 & 48.45 & 43.75 & 44.34 \\
\textbf{Nuextract 2.0 8B}     & 52.53 & 68.30 & 54.39 & 52.07 & 52.09 & 62.87 & 57.01 & 57.92 & 54.67 \\
\textbf{Qwen2.5 3B Instruct}  & \textbf{65.75} & 55.20 & 49.21 & 53.32 & 59.15 & 59.11 & 57.46 & 59.47 & 53.05 \\
\textbf{MeXtract 0.5B}    & 62.39 & 77.08 & 66.67 & 67.01 & 71.47 & 70.55 & 70.07 & 74.17 & 64.66 \\
\textbf{MeXtract 1.5B}    & 64.66 & \textbf{80.52} & 71.59 & 74.78 & 72.50 & 74.67 & 73.10 & 75.62 & 68.49 \\
\textbf{MeXtract 3B}      & 61.49 & 76.01 & \textbf{74.24} & \textbf{76.09} & \textbf{76.25} & \textbf{76.45} & \textbf{76.83} & \textbf{76.12} & \textbf{71.90} \\
\bottomrule
\end{tabular}
\end{table*}

\begin{figure*}[h!]
\centering
\begin{tcolorbox}[colback=gray!5,colframe=black!40,width=0.9\linewidth]
\small
\texttt{You are an AI assistant that classifies papers into multiple categories using the title and abstract of the paper.
You should predict if the paper introduces or releases a new dataset or benchmark for nlp, computer vision or speech.
The abstract MUST explicitly mention the creation of a new dataset. Do NOT make any assumptions.
Each dataset paper must be classified in one of the following categories:
- ar: the paper introduces a new  dataset in Arabic or its dialects
- en: the paper introduces a new dataset in English, if the language is not mentioned then assume it is English.
- fr: the paper introduces a new dataset in French
- ru: the paper introduces a new dataset in Russian
- jp: the paper introduces a new dataset in Japanese
- other: the paper introduces a new dataset in another language not in the list above
- multi: the paper introduces a new multilingual or cross-lingual dataset
If the paper is doesn't belong to any of the categories above, the category should be "none".
The output should be a JSON object with the following fields:
{
    "reasoning": "reasoning why the paper is in the category",
    "category": "one item from the list [ar, en, fr, ru, jp, other, multi, none]"
}}
\end{tcolorbox}
\caption{System prompt used to extract resource papers. We use Gemma 3 27B to label datasets into 8 categories.}
\label{fig:system_prompt}
\end{figure*}

\onecolumn
\begin{lstlisting}[language=Python, caption=Model Schema., breaklines=true, numbers=none, label=lst:schema]
{
    "Name": {
        "answer_type": "str",
        "answer_min": 1,
        "answer_max": 5
    },
    "Num_Parameters": {
        "answer_type": "float",
        "answer_min": 1,
        "answer_max": 1000
    },
    "Unit": {
        "answer_type": "str",
        "answer_min": 1,
        "answer_max": 1,
        "options": [
            "Million",
            "Billion",
            "Trillion"
        ]
    },
    "Type": {
        "answer_type": "str",
        "answer_min": 1,
        "answer_max": 1,
        "options": [
            "Base",
            "Code",
            "Chat"
        ]
    },
    "Think": {
        "answer_type": "bool",
        "answer_min": 1,
        "answer_max": 1
    },
    "Version": {
        "answer_type": "float",
        "answer_min": 0.0
    },
    "Models": {
        "answer_type": "list[dict[Name, Num_Parameters, Unit, Type, Think]]",
        "answer_min": 1,
        "answer_max": 10
    },
    "License": {
        "answer_type": "str",
        "answer_min": 1,
        "answer_max": 1,
        "options": [
            "Apache-1.0",
            "Apache-2.0",
            "Non Commercial Use - ELRA END USER",
            "BSD",
            "CC BY 1.0",
            "CC BY 2.0",
            "CC BY 3.0",
            "CC BY 4.0",
            "CC BY-NC 1.0",
            "CC BY-NC 2.0",
            "CC BY-NC 3.0",
            "CC BY-NC 4.0",
            "CC BY-NC-ND 1.0",
            "CC BY-NC-ND 2.0",
            "CC BY-NC-ND 3.0",
            "CC BY-NC-ND 4.0",
            "CC BY-SA 1.0",
            "CC BY-SA 2.0",
            "CC BY-SA 3.0",
            "CC BY-SA 4.0",
            "CC BY-NC 1.0",
            "CC BY-NC 2.0",
            "CC BY-NC 3.0",
            "CC BY-NC-SA 1.0",
            "CC BY-NC-SA 2.0",
            "CC BY-NC-SA 3.0",
            "CC BY-NC-SA 4.0",
            "CC BY-NC 4.0",
            "CC0",
            "CDLA-Permissive-1.0",
            "CDLA-Permissive-2.0",
            "GPL-1.0",
            "GPL-2.0",
            "GPL-3.0",
            "LDC User Agreement",
            "LGPL-2.0",
            "LGPL-3.0",
            "MIT License",
            "ODbl-1.0",
            "MPL-1.0",
            "MPL-2.0",
            "ODC-By",
            "AFL-3.0",
            "CDLA-SHARING-1.0",
            "unknown",
            "custom"
        ]
    },
    "Year": {
        "answer_type": "year",
        "answer_min": 1900,
        "answer_max": 2025
    },
    "Benchmarks": {
        "answer_type": "list",
        "answer_min": 1,
        "answer_max": 64
    },
    "Architecture": {
        "answer_type": "str",
        "answer_min": 1,
        "answer_max": 1,
        "options": [
            "Transformer",
            "MoE",
            "SSM",
            "RNN",
            "CNN",
            "Hybrid",
            "other"
        ]
    },
    "Context": {
        "answer_type": "int",
        "answer_min": 1
    },
    "Language": {
        "answer_type": "str",
        "answer_min": 1,
        "answer_max": 1,
        "options": [
            "monolingual",
            "bilingual",
            "multilingual"
        ]
    },
    "Provider": {
        "answer_type": "str",
        "answer_min": 1,
        "answer_max": 5
    },
    "Modality": {
        "answer_type": "str",
        "answer_min": 1,
        "answer_max": 1,
        "options": [
            "text",
            "audio",
            "video",
            "image",
            "multimodal"
        ]
    },
    "Paper_Link": {
        "answer_type": "url",
        "answer_min": 1,
        "answer_max": 1
    }
}
\end{lstlisting}

\end{document}